\documentclass[runningheads]{llncs}
\usepackage[T1]{fontenc}
\usepackage{float}

\usepackage{tabularx}
\usepackage{algorithm}
\usepackage{algorithmic} 
\usepackage{url}
\usepackage{booktabs}
\usepackage{xcolor}
\usepackage{subcaption}
\usepackage{amssymb}
\usepackage{multirow}
\usepackage{amsmath}
\usepackage{makecell}

\usepackage{graphicx}
\begin{document}
\newcommand{\sukhi}{\textcolor{blue}}

\title{Fast Tensorization of Neural Networks via Slice-wise Feature Distillation} 
%
%
\author{
Safa Hamreras\inst{1} \and
Sukhbinder Singh\inst{2} \and
Rom\'an Or\'us\inst{1,3,4}
}

\authorrunning{S. Hamreras et al.}

\institute{
Donostia International Physics Center, Paseo Manuel de Lardizabal 4, E-20018 San Sebasti\'an, Spain
\email{safa.hamreras@dipc.org}
\and
Multiverse Computing, Spadina Ave., Toronto, ON M5T 2C2, Canada
\email{sukhi.singh@multiversecomputing.com}
\and
Multiverse Computing, Paseo de Miram\'on 170, E-20014 San Sebasti\'an, Spain
\and
Ikerbasque Foundation for Science, Maria Diaz de Haro 3, E-48013 Bilbao, Spain
\email{roman.orus@multiversecomputing.com}
}
\maketitle              
\begin{abstract}
We propose a scalable tensorization framework for neural network compression based on slice-wise feature distillation. Unlike conventional tensor decomposition methods that rely on costly global fine-tuning, our approach decomposes the network into slices consisting of either individual layers or blocks (e.g., convolutional layers or MLPs), or \textit{small} groups of \textit{consecutive} layers, and tensorizes each slice \textit{independently} to reproduce the intermediate representations of the original pretrained model. This modular strategy improves accuracy recovery, reduces data requirements, and enables efficient parallel optimization. Experiments on ResNet-34 show significant gains over conventional global tensorization, achieving near-lossless compression at moderate compression rates with faster optimization. Results on GPT-2 XL further demonstrate the scalability of the method and its applicability to large-scale models, particularly in distributed settings. 

\textbf{Keywords:} Neural Network Compression, Tensor Networks, Feature Distillation.
\end{abstract}
\section{Introduction}

Tensor Networks (TNs) have emerged as a quantum-inspired framework for neural network compression \cite{novikov2015tensorizing}. Tensorization consists of factorizing large weight matrices into compact, low-rank tensor networks (such as Matrix Product Operators \cite{schollwock2011density} or Tucker tensors \cite{tucker1966some}) that exploit inherent low-rank structure in the weights. While this approach has shown promise \cite{ma2019tensorized,liu2022deep,xu2023tensorgpt,tomut2024compactifai}, it remains underexplored compared to pruning and quantization \cite{hamreras2025tensorization}.

A major limitation of standard tensorization pipelines is that they rely on direct weight approximation (e.g., via SVD-based initialization \cite{klema1980singular}), which often leads to substantial degradation in model performance. Since minimizing matrix reconstruction error does not necessarily preserve the network’s functional behavior \cite{hsu2022language}, a costly global end-to-end fine-tuning stage is typically required \cite{tomut2024compactifai,lebedev2014speeding,liu2021enabling}. Meanwhile, the success of pruning and quantization has shown that matching internal features, rather than just weights, plays a crucial role in maintaining accuracy \cite{lin2024awq,jiao2019tinybert}.

In this paper, we introduce a \emph{slice-wise feature distillation} strategy to address these limitations. Our approach partitions the network into modular slices (ranging from single layers to larger contiguous blocks) and applies tensor decomposition to each slice. Then, instead of performing global fine-tuning, we independently optimize each slice to reproduce the output activations of the corresponding original slice. This local healing is performed using feature distillation, where inputs are fed to the tensorized slice and its outputs are trained to match those of the original model via a simple MSE objective. Since slices are optimized independently, this strategy eliminates the need to keep the full model in memory during training, and it provides several key advantages:
\begin{itemize}
    \item \textbf{Stronger performance recovery} through richer, slice-local supervision signals compared to global task-level fine-tuning.
    \item \textbf{Faster and more scalable optimization}, as the problem is broken into smaller, independent subproblems that can be solved in parallel and asynchronously.
    \item \textbf{Improved data efficiency}, since optimization is based on intermediate features at several levels of the model, providing rich information.
    \item \textbf{Effective initialization for global fine-tuning}, where slice-wise tensorization followed by global finetuning lead to better results at higher compression rates, as demonstrates some of our experiments.
\end{itemize}

The remainder of the paper is organized as follows: In In Sec.~\ref{work}, we review the related work and position our contribution within this landscape. In Sec.\ref{sec:meth}, we explain our methodology of feature distillation-based tensorization, in Sec.\ref{sec:experiments}, we report our results and discuss them, and we  conclude with a summary of our findings and a discussion of our future perspectives in Sec.\ref{conclusions}.  

\section{Related work}\label{work}

In this section, we review prior work most relevant to our tensorization strategy, covering neural network tensorization, knowledge distillation, and activation-aware compression.

\textbf{Neural network tensorization.} Tensorization has emerged as a flexible and effective neural network compression strategy, where dense weight matrices are factorized into compact tensor network representations. This approach has been successfully applied across a wide range of architectures. In large language models (LLMs), tensorization has been explored at different architectural levels: TensorGPT (\cite{xu2023tensorgpt}) tensorizes GPT-2 embeddings using Matrix Product States (MPS) (\cite{perez2006matrix}), while compactifAI (\cite{tomut2024compactifai}) replaces self-attention and multi-layer perceptron layers with Matrix Product Operator (MPO) layers. In convolutional neural networks, tensorization is commonly performed using Tucker decomposition (\cite{wang2024tucker,liu2022deep}) or Canonical Polyadic (CP) decomposition (\cite{lebedev2014speeding,astrid2017cp}). Other applications include transformers (\cite{liu2023efficient}), graph neural networks (\cite{hua2022high}), and quantum neural networks (\cite{liu2024tensor}).

However, many existing tensorization methods rely on global end-to-end finetuning after decomposition to recover performance, which can become computationally expensive for large tensorized models, may impose substantial memory overhead, and limits practical scalability. Our work addresses this limitation through localized, feature-distillation-based recovery over independently optimized model slices.

\textbf{Neural network distillation.}
Knowledge distillation has been widely used for neural network compression (\cite{jiao2019tinybert,polino2018model}). In this setting, a compact \textit{student} model is trained to mimic the behavior of a larger \textit{teacher} model, often achieving competitive performance with significantly fewer parameters (\cite{gou2021knowledge}). Distillation methods are commonly categorized into: (1) response-based distillation, where the student matches the final teacher outputs (\cite{hinton2015distilling}); (2) feature-based distillation, where intermediate representations are also matched (\cite{jiao2019tinybert}); and (3) relation-based distillation, which captures structural relationships between layers or samples (\cite{passalis2020heterogeneous,gou2021knowledge}).

Our method draws inspiration from feature-based distillation, as the tensorized model is trained to reproduce intermediate features of the pretrained model after tensor decomposition. A representative example is TinyBERT (\cite{jiao2019tinybert}), where a smaller student model is trained through layer-wise distillation to mimic intermediate representations of BERT (\cite{devlin2019bert}). Unlike these distillation methods, which train a separate compact architecture, our method preserves the original architecture and uses feature distillation specifically to recover performance after tensor decomposition. This post-decomposition local healing is a key distinction of our approach.

\textbf{Activation-aware compression.}
Several compression methods aim to preserve the intermediate activations of the original  model, a strategy that has proven highly effective in quantization (\cite{chen2019deep,yao2022zeroquant,frantar2022optimal,frantaroptq,kuzmin2023pruning}) and low-rank compression. In these approaches, compressed weights are optimized such that the output activations remain close to those of the pretrained model. Among low-rank methods, ASVD (\cite{yuan2023asvd}) and FWSVD (\cite{hsu2022language}) explicitly optimize SVD-based approximations to preserve original layer activations. ASVD uses activation statistics to improve low-rank approximations, while FWSVD incorporates Fisher-weighted importance scores during truncation. Similarly, SVD-LLM (\cite{wang2024svd}) uses activation-aware whitening and layer-wise closed-form updates to better align compressed outputs with their pre-tensorized counterparts.

While our method shares the activation-preserving objective of these approaches, it differs fundamentally in optimization strategy. Prior activation-aware methods typically improve compressed layers through analytical reconstruction or local approximation objectives. In contrast, our method treats tensorization recovery as a feature-distillation problem over independently optimized model slices, which can range from individual layers to larger contiguous sub-networks, enabling modular optimization, flexible granularity, and natural parallelism.

\section{Methodology --- Tensorization via Feature Distillation}
\label{sec:meth}

A neural network with $L$ layers can be partitioned into $N$ modular slices, each having $M$ layers, such that $L = N \times M$. Our idea consists of tensorizing the layers (e.g., linear or convolutional) in each slice of the neural network independently, where $M=1$ corresponds to layerwise tensorization. In this setup, each slice---after applying tensor decomposition---aims to reproduce the output features of the corresponding pretrained slice in a student-teacher fashion. 

\begin{figure}[H]
    \centering
    \includegraphics[width=.85\columnwidth]{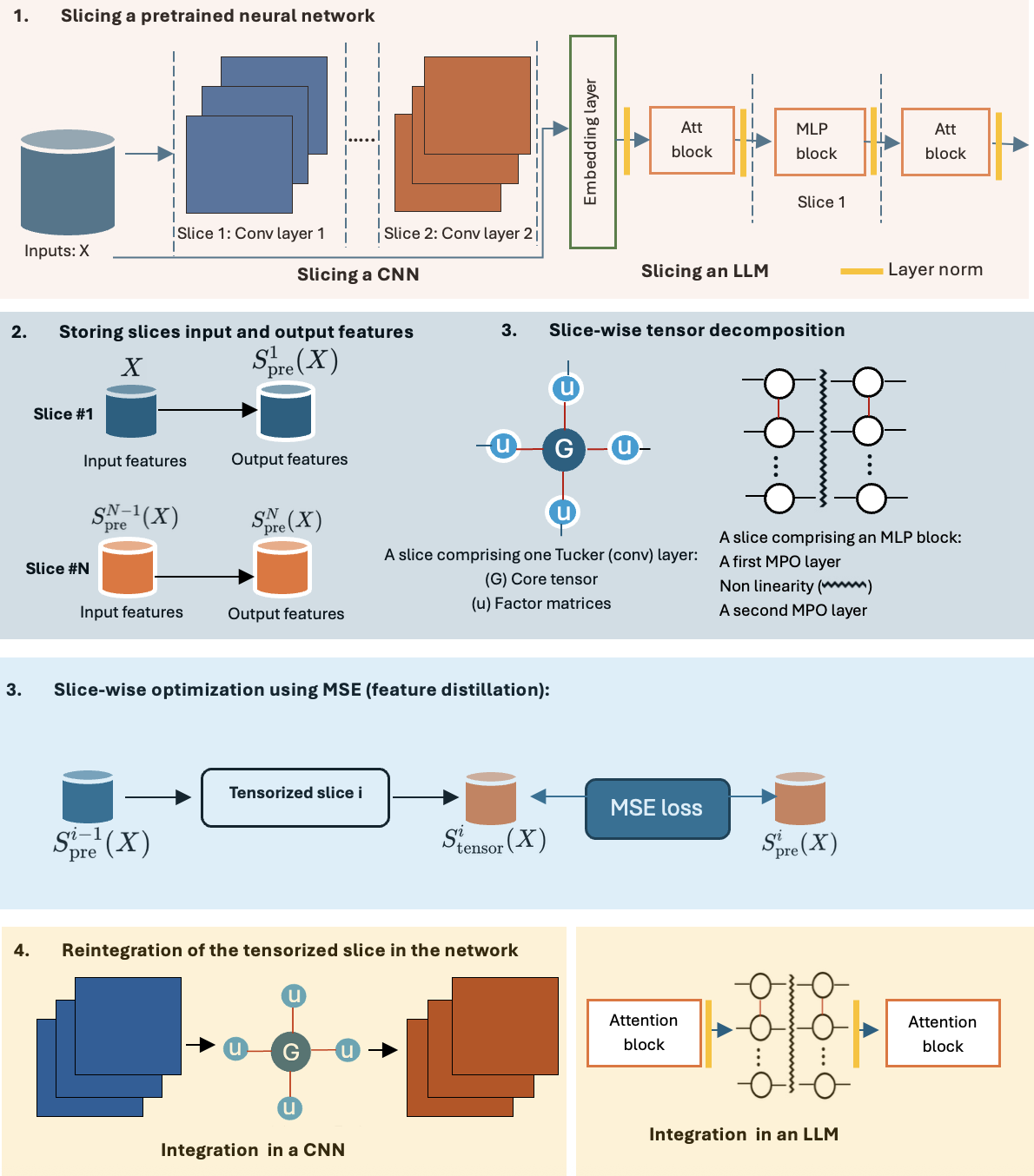}
    \caption{The process of slice-wise feature distillation: (1) First split a pretrained neural network into slices, where one slice can include a convolutional layer/ MLP block, (2) Training data is passed through the pretrained network and input and output features for the slice to tensorize are stored, (3) Each layer within the slice to tensorize is decomposed into Tucker/MPO layer, depending on its type, (4) Tucker and MPO slices are optimized by feature-based distillation using MSE, (5) Finally, the optimized slices are reintegrated again into the neural network.}
\end{figure}

That says, given a pretrained  model, the training data $X$ is first passed through the model spanning the first layer until the slice $S^{[i]}$, which is the slice to tensorize. The input and output features of this slice can then be recorded. We refer to the input and output features of the pretrained slice $S^{[i]}$ as $S^{[i-1]}_{\text{pre}}(X)$ and $S^{[i]}_{\text{pre}}(X)$, respectively, while the output features of the tensorized slice are referred to as $S^{[i]}_{\text{tensor}}(X)$.

Given a slice $S^{[i]}$ to tensorize, this latter can be detached from the neural network and the corresponding  weight matrices are decomposed based on MPO or Tucker decomposition. After that, this slice is fed the  input features of the corresponding  slice in the pretrained model, that is $S^{[i-1]}_{\text{pre}}(X)$, with the aim to reproduce its  output features $S^{[i]}_{\text{pre}}(X)$. As such, in order to tensorize one slice, Once pretrained intermediate features are collected, optimization can be performed without repeatedly executing the full model. when the training is done, the tensorized slice is plugged back into the  network.

The objective function used for training is the Mean Squared Error (MSE) between the tensorized slice output and its corresponding pretrained slice output:

\begin{equation}
\label{eq:aoloss}
\mathrm{MSE} =
\frac{1}{N}
\left\|
S^{[i]}_{\text{tensor}}(X)
-
S^{[i]}_{\text{pre}}(X)
\right\|_F^2
\end{equation}

Where $N$ is the number of data points (features) and $i$ is the slice index.


\section{Experiments and Results}
\label{sec:experiments}

We evaluate the proposed slice-wise feature-distillation based tensorization on both convolutional neural networks (CNNs) and large language models (LLMs) to assess its effectiveness and scalability across architectures. Throughout this section and for simplicity reasons, we refer to slice-wise tensorization based on feature distillation as \textbf{local tensorization}, while conventional end-to-end finetuning of the tensorized model using task supervision is referred to as \textbf{global tensorization}.

Local tensorization requires storing intermediate features from the pretrained model, introducing a one-time I/O overhead that depends on the hardware and implementation. Since these features are computed once and reused across training iterations, this cost is amortized over optimization and can also be overlapped with training. To isolate the algorithmic efficiency of the optimization itself, reported optimization times for both methods include only forward and backward computation, excluding data loading and feature extraction overhead.

\subsection{CNN Experiments: ResNet-34}

All CNN experiments are conducted on an AWS \texttt{g5.8xlarge} instance equipped with 32 vCPUs, 128~GiB RAM, and one NVIDIA A10G Tensor Core GPU with 24~GiB VRAM.

We evaluate our method on ResNet-34 \cite{he2016deep} using the CIFAR-10 and CIFAR-100 datasets\footnote{\url{https://www.cs.toronto.edu/~kriz/cifar.html}}. Tensorization is performed using Tucker decomposition at two compression rates (CR): 0.5 and 0.7. We define compression rate as:

\[
\mathrm{CR} =
\frac{\text{Parameter count of original model} - \text{Parameter count of tensorized model}}
{\text{Parameter count of original model}}.
\]

To construct the compressed model, we first profile the convolutional layers to identify those particularly sensitive to tensorization (see Appendix). That says, we tensorize all $3 \times 3$ convolutional layers, excluding 4 layers.

Specifically, a convolutional layer with weight tensor shape $[s_1, s_2, s_3, s_4]$ is replaced by a Tucker-decomposed representation with ranks $[r_1, r_2, s_3, s_4]$, where $r_1 < s_1$ and $r_2 < s_2$. The ranks are selected to achieve the desired compression rate. Tucker decomposition is implemented using the TensorLy library\footnote{\url{https://tensorly.org/stable/index.html}}, with randomly initialized tensor factors.

Local and global tensorization are independently hyperparameter-tuned at CR = 0.5 for each comparison to ensure a fair best-performance comparison (see Appendix), and the same configuration is reused at CR = 0.7.

\subsection{Experiments on CIFAR-10}

Local tensorization uses the Adam optimizer with a batch size of 8 and a learning rate of 0.001, while global tensorization uses Adam with a batch size of 16 and a learning rate of 0.0005.

We first evaluate tensorization at a moderate CR of 0.5. The tensorized ResNet-34 is optimized using either local or global tensorization, and we additionally investigate the effect of reducing the training data available for local tensorization. Specifically, local tensorization is evaluated by fine-tuning on training subsets of 10k, 30k, and 50k samples, while global tensorization is evaluated when finetuned on the complete 50k training set.

\begin{figure}[!ht]
    \centering

    \begin{subfigure}{0.49\textwidth}
        \centering
        \includegraphics[width=\linewidth]{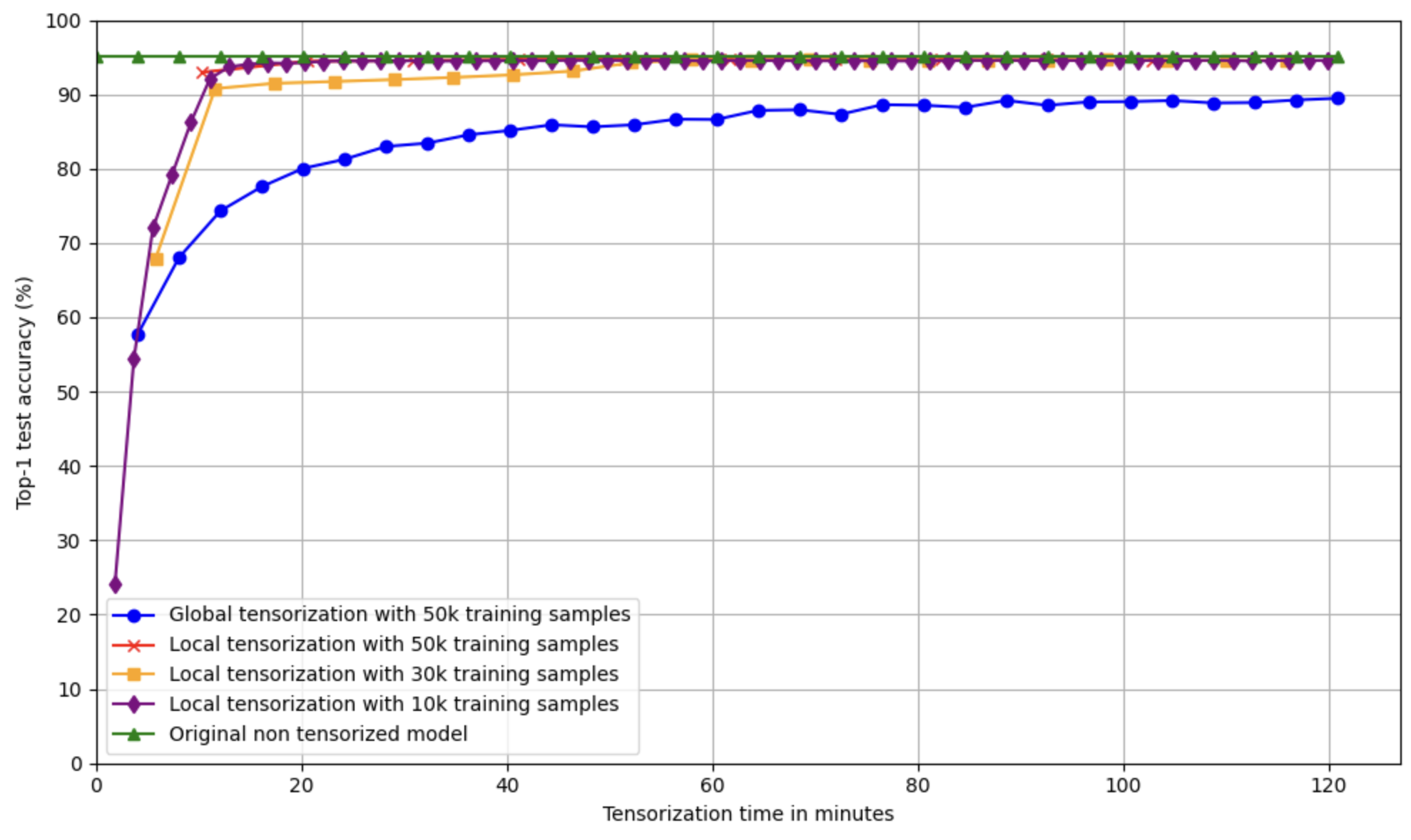}
        \caption{Cifar-10, CR = 0.5}
        \label{fig:Cifar10-cr05}
    \end{subfigure}%
    \begin{subfigure}{0.49\textwidth}
        \centering
        \includegraphics[width=\linewidth]{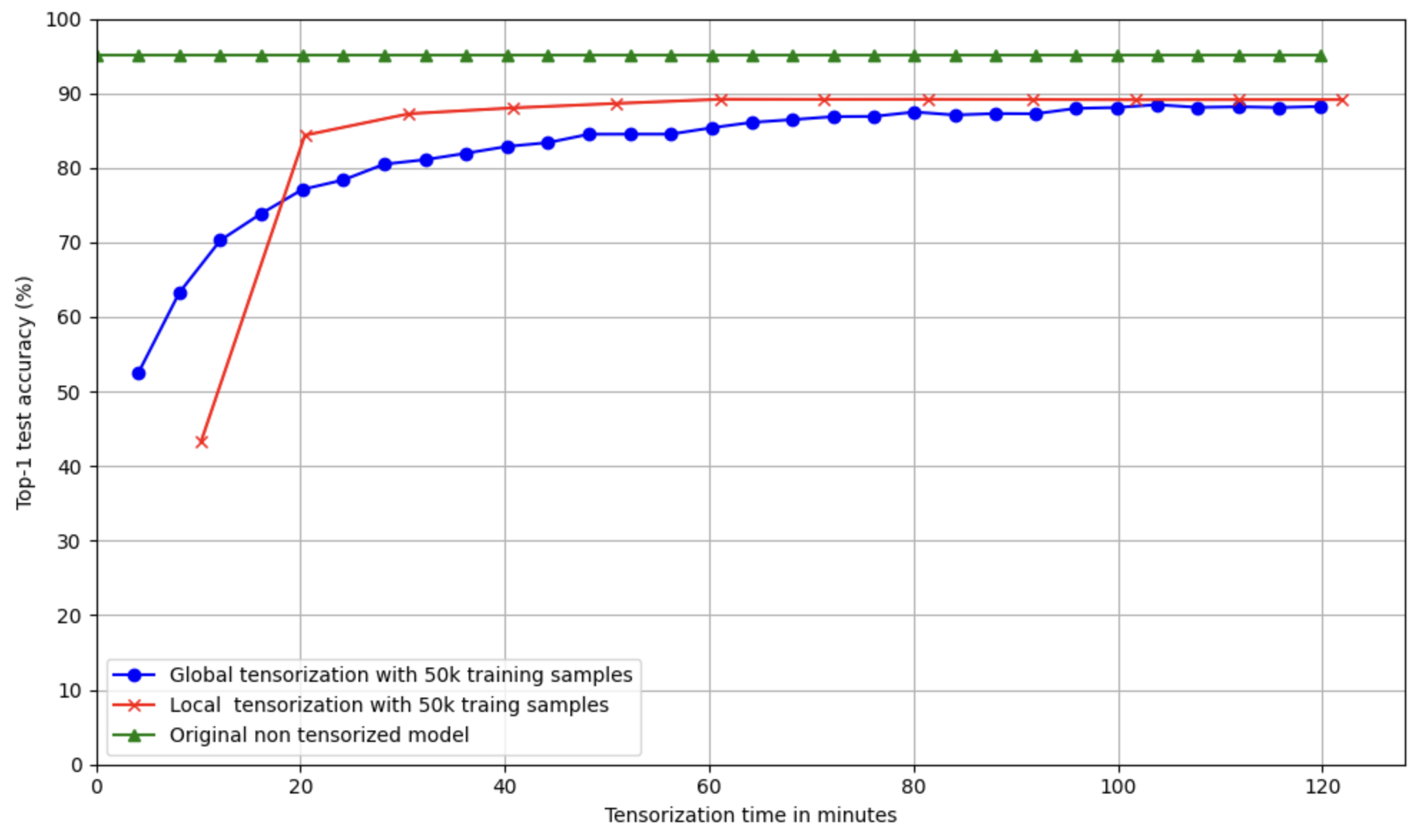}
        \caption{Cifar-10, CR = 0.7}
        \label{fig:Cifar10-cr07}
    \end{subfigure}

    \vspace{0.2cm}

    \begin{subfigure}{0.49\textwidth}
        \centering
        \includegraphics[width=\linewidth]{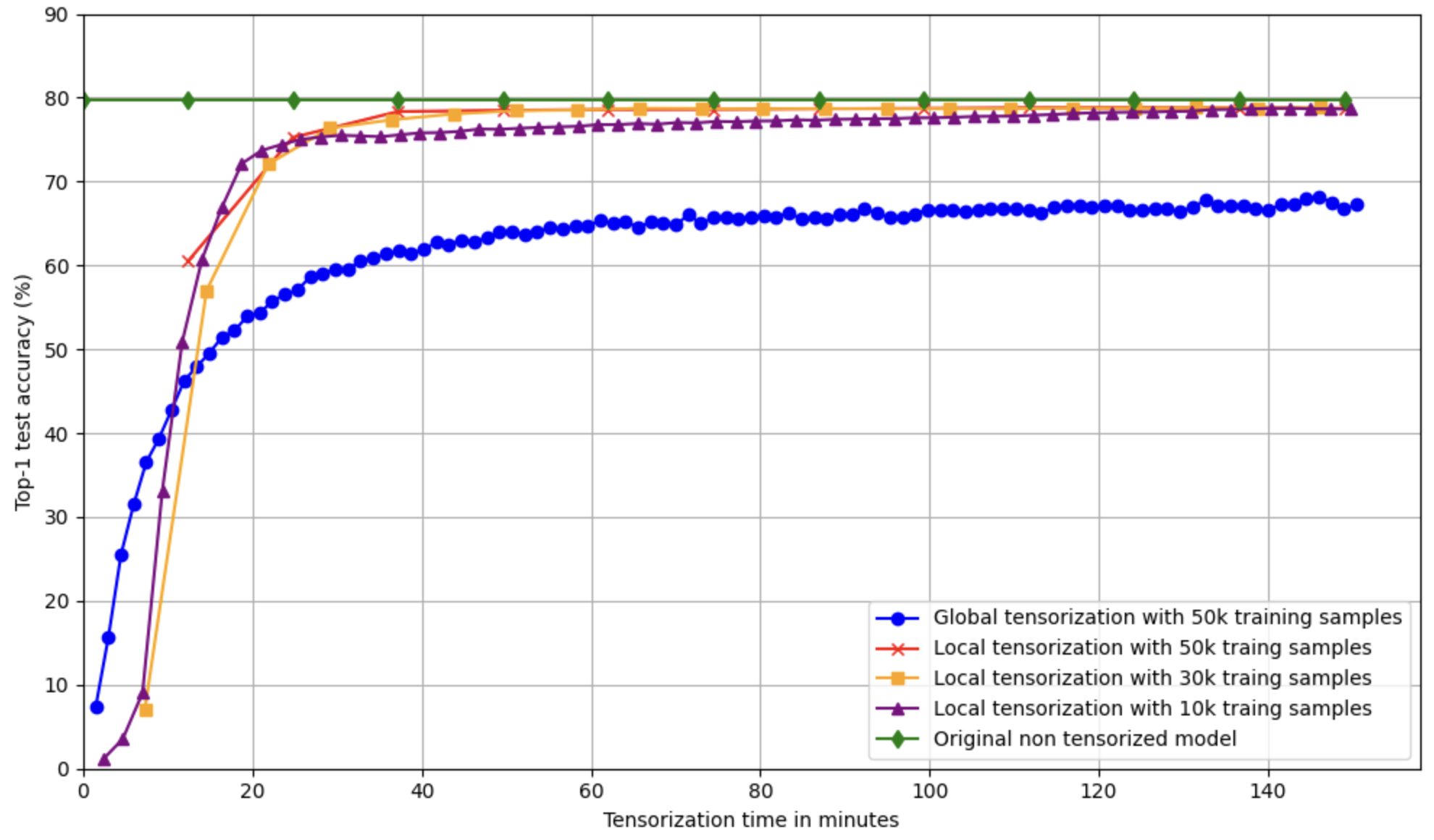}
        \caption{Cifar-100, CR = 0.5}
        \label{fig:Cifar100-cr05}
    \end{subfigure}%
    \begin{subfigure}{0.49\textwidth}
        \centering
        \includegraphics[width=\linewidth]{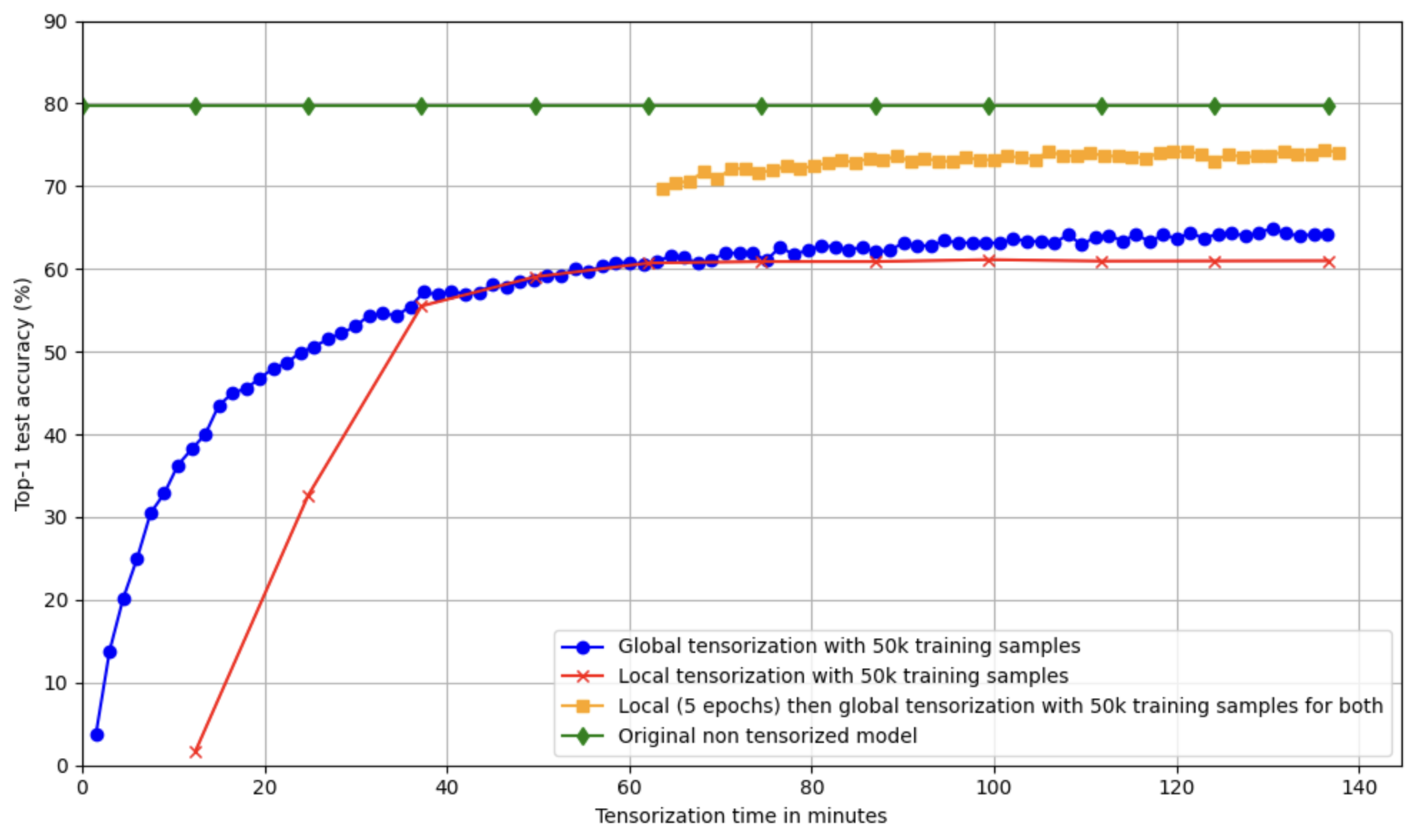}
        \caption{Cifar-100, CR = 0.7}
        \label{fig:Cifar100-cr07}
    \end{subfigure}

    \caption{
     Comparison between local and global tensorization of 3×3 convolutional layers
    using Tucker decomposition on Resnet-34. All convolutional layers are tensorized except layers with indices 85, 55, 29, and 93,
    which are excluded due to their sensitivity, and in order to achieve the targeted compression rate. Local tensorization optimizes layers independently by matching pretrained model outputs, while global tensorization freezes non-tensorized layers and performs end to end finetuning. 
    Markers indicate training epochs for local and global tensorization and Top-1 test accuracy is reported. }
    \label{fig:tensorization-comparison}
\end{figure}

Figure~\ref{fig:Cifar10-cr05} shows that local tensorization consistently converges faster and reaches higher final accuracy than global tensorization across all data regimes. Even when using only 10k samples, local tensorization remains highly effective, indicating strong data efficiency.

Table~\ref{table-datasets-Cifar10} summarizes the best Top-1 accuracies. Local tensorization with 50k samples achieves the highest performance (94.70\%), recovering over 99\% of the original pretrained model accuracy (95.04\%) and outperforming global tensorization by +5.23\%. Notably, reducing the local tensorization dataset to 30k or even 10k samples leads only to marginal performance degradation, further supporting the robustness of local tensorization.

\begin{table}[htbp]
\centering
\caption{Top-1 test accuracy (\%) comparison on CIFAR-10 for CR = 0.5. The labels 50k/30k/10k denote the number of training images used during tensorization.}
\label{table-datasets-Cifar10}
\begin{tabularx}{\textwidth}{lXXXXX}
\toprule
\textbf{} 
& \textbf{ResNet-34} 
& \textbf{Tensorized Local 50k} 
& \textbf{Tensorized Local 30k} 
& \textbf{Tensorized Local 10k} 
& \textbf{Tensorized Global 50k} \\
\midrule
Accuracy & 95.04 & 94.70 & 94.69 & 94.61 & 89.47 \\
\bottomrule
\end{tabularx}
\end{table}

We next increase the compression rate to  0.7 to evaluate a more aggressive compression regime. As shown in Figure~\ref{fig:Cifar10-cr07}, global tensorization initially improves faster during the first optimization steps, but local tensorization rapidly catches up and ultimately achieves slightly higher final Top-1 accuracy.

Table~\ref{tab:cifar10_comparison}  summarizes the performance across both compression rates. At CR = 0.5, local tensorization clearly outperforms global tensorization in both accuracy and optimization speed. At CR = 0.7, the performance gap narrows, but local tensorization remains competitive while preserving its efficiency advantage.

\begin{table}[htbp]
\centering
\caption{Comparison of local and global tensorization on CIFAR-10 in terms of accuracy (\%) and optimization time (minutes) at different compression rates.}
\label{tab:cifar10_comparison}
\begin{tabular}{lcccccc}
\toprule
\multirow{2}{*}{\textbf{Method}} 
& \multicolumn{3}{c}{\textbf{CR = 0.5}} 
& \multicolumn{3}{c}{\textbf{CR = 0.7}} \\
\cmidrule(lr){2-4} \cmidrule(lr){5-7}
& \textbf{Top-1} & \textbf{Top-5} & \textbf{Time}
& \textbf{Top-1} & \textbf{Top-5} & \textbf{Time} \\
\midrule
Local  & \textbf{94.70} & \textbf{99.84} & \textbf{51.38}
       & \textbf{89.19} & 99.51 & \textbf{61.11} \\
Global & 89.47 & 99.49 & 120.88
       & 88.46 & \textbf{99.54} & 103.81 \\
\bottomrule
\end{tabular}
\end{table}

To quantify optimization efficiency, Table~\ref{tab:cifar10_comparison} also reports the  optimization time required to reach the best Top-1 accuracy for each method. At CR = 0.5, local tensorization achieves a 2.35$\times$ speedup over global tensorization, while at CR = 0.7 the speedup remains substantial at 1.7$\times$, demonstrating that the proposed method maintains high performance even under more aggressive compression rates.

\subsection{Experiments on CIFAR-100}

For CIFAR-100, local tensorization uses the Adam optimizer with a batch size of 8 and a learning rate of 0.0005, while global tensorization uses Adam with a batch size of 64 and the same learning rate of 0.0005.

We first evaluate tensorization at CR = 0.5, using the same experimental protocol as in CIFAR-10. 

Figure~\ref{fig:Cifar100-cr05} shows that local tensorization consistently converges faster than global tensorization across all data regimes. Moreover, local tensorization nearly recovers the original pretrained model performance despite compression. Table~\ref{table:Cifar100-datasets} reports the best Top-1 accuracies: local tensorization with 50k samples achieves 78.81\%, compared to 79.79\% for the pretrained model, recovering approximately 98.8\% of the original performance while outperforming global tensorization by more than 10\%.

As in CIFAR-10, reducing the local tensorization dataset size has only a minor impact on final performance, indicating strong data efficiency at a more complex task.

\begin{table}[htbp]
\centering
\caption{Top-1 test accuracy (\%) comparison on CIFAR-100 for CR = 0.5. The labels 50k/30k/10k denote the number of training images used during tensorization.}
\label{table:Cifar100-datasets}
\begin{tabularx}{\textwidth}{lXXXXX}
\toprule
\textbf{} 
& \textbf{ResNet-34} 
& \textbf{Tensorized Local 50k} 
& \textbf{Tensorized Local 30k} 
& \textbf{Tensorized Local 10k} 
& \textbf{Tensorized Global 50k} \\
\midrule
Accuracy & 79.79 & 78.81 & 78.80 & 78.74 & 68.19 \\
\bottomrule
\end{tabularx}
\end{table}

We next evaluate a more aggressive compression setting at CR = 0.7. In this regime, global tensorization achieves stronger final performance than local tensorization alone, suggesting that highly compressed models may benefit from some degree of end-to-end optimization.

To address this limitation, we introduce a hybrid \textbf{local + global} strategy: the model is first optimized locally for five epochs using local tensorization, then further refined using global end-to-end finetuning. In this setting, local tensorization serves as an initialization stage for global recovery.

This hybrid strategy substantially improves performance, achieving 74.22\% Top-1 accuracy compared to 65.12\% for global tensorization alone, while requiring less optimization time than full global recovery. This result suggests that feature-distillation-based tensorization provides a stronger initialization for highly compressed tensorized models.

Tables~\ref{table-Cifar100-results} summarize the full CIFAR-100 results.

\begin{table}[htbp]
\centering
\caption{CIFAR-100 results for Local, Global, and Hybrid (Local + Global) tensorization at different compression rates.}
\label{table-Cifar100-results}

\small   

\begin{tabular}{lcccccc}
\toprule
\multirow{2}{*}{\textbf{Method}} 
& \multicolumn{2}{c}{\textbf{CR = 0.5}} 
& \multicolumn{3}{c}{\textbf{CR = 0.7}} \\
\cmidrule(lr){2-3} \cmidrule(lr){4-6}
& \textbf{Top-1} 
& \textbf{Time}
& \textbf{Top-1}
& \textbf{Top-5}
& \textbf{Time} \\
\midrule
Local          & 78.81 & \textbf{124.12} & 61.12 & 85.94 & \textbf{99.36} \\
Global         & 68.19 & 145.99           & 65.12 & 87.73 & 144.08 \\
Local + Global & --    & --               & \textbf{74.22} & \textbf{92.29} & 106.01 \\
\bottomrule
\end{tabular}
\end{table}

\subsection{Comparison with Prior Work}

To position our method within the broader neural network compression literature, Table~\ref{table-comp-cifar} compares our approach against representative prior methods on ResNet-34 for CIFAR-10 and CIFAR-100. Since most of methods rely on different compression paradigms (e.g., pruning, structured compression, low-rank decomposition), the comparison is intended to provide context rather than a strict like-for-like evaluation.

The compared methods include APSSF \cite{geng2024apssf}, which uses adaptive filter clustering and pruning; the method in \cite{wei2022automatic}, based on reinforcement learning for automatic structured pruning; NC-CTD \cite{sun2020deep}, which exploits inter-layer redundancy through coupled tensor decomposition; LJSVD \cite{chen2023joint}, based on joint low-rank decomposition across layers; and Edropout \cite{salehinejad2021edropout}, which performs mask-based pruning of filters and neurons. For fairness, the reported accuracy change for each method is computed relative to the original pretrained model accuracy reported in its respective paper.

At moderate compression (CR = 0.5), our method exhibits only minor performance degradation on both datasets, with a Top-1 accuracy drop of 0.34\% on CIFAR-10 and 0.98\% on CIFAR-100. Some pruning-based methods achieve slightly stronger accuracy retention at comparable compression levels, reflecting the relative maturity of pruning as a compression paradigm. At higher compression rates, larger performance degradation is observed, as expected. In this regime, approaches specifically designed for aggressive structured compression or joint low-rank optimization may retain stronger accuracy. However, these methods operate under different compression assumptions and optimization strategies.

Importantly, our objective is not to outperform all compression methods in  accuracy, but to improve tensorization as a scalable and efficient compression framework. Compared to standard tensorization pipelines, our results show substantially stronger recovery through feature-distillation-based local optimization, while preserving the modularity and flexibility of tensor decompositions.

Tensorization is also complementary to other compression strategies such as pruning and quantization, suggesting that additional gains may be achievable through hybrid compression pipelines.

\begin{table}[t]
\centering
\caption{Top-1 accuracy change (compressed vs.\ original, \%) of ResNet-34 compared with representative prior methods. For fairness, the reported accuracy change for each method is computed relative to the original model accuracy reported in its respective paper.}
\label{table-comp-cifar}

\begin{subtable}[t]{0.48\linewidth}
\centering
\caption{CIFAR-10}
\label{Cifar10-compare}
\begin{tabular}{lcccc}
\toprule
\textbf{Method} & \textbf{0.5} & \textbf{0.7} & \textbf{0.32} & \textbf{0.95} \\
\midrule
Ours                        & -0.34  & -5.85 & --    & --    \\
APSSF \cite{geng2024apssf} & +0.02  & --    & --    & --    \\
Method in \cite{wei2022automatic}    & +0.19  & -0.8  & --    & --    \\
NC-CTD \cite{sun2020deep}  & --     & --    & +1.77 & --    \\
LJSVD \cite{chen2023joint} & --     & --    & --    & -1.14 \\
\bottomrule
\end{tabular}
\end{subtable}
\hfill
\begin{subtable}[t]{0.48\linewidth}
\centering
\caption{CIFAR-100}
\label{Cifar100-compare}
\begin{tabular}{lcccc}
\toprule
\textbf{Method} & \textbf{0.5} & \textbf{0.7} & \textbf{0.3} & \textbf{0.95} \\
\midrule
Ours                                     & -0.98   & -5.57 & --     & --    \\
Edropout \cite{salehinejad2021edropout} & -5.17   & --    & --     & --    \\
APSSF \cite{geng2024apssf}              & +0.003  & --    & --     & --    \\
NC-CTD \cite{sun2020deep}               & --      & --    & +11.97 & --    \\
LJSVD \cite{chen2023joint}              & --      & --    & --     & -1.42 \\
\bottomrule
\end{tabular}
\end{subtable}

\end{table}

\subsection{Experiments on GPT-2 XL}

\subsubsection{Experimental Settings}

To evaluate scalability on large language models, we conduct experiments on GPT-2 XL \cite{radford2019language} using a system equipped with one NVIDIA H100 GPU, 400 GB RAM, and a 100-core CPU.

Training is performed on a subset of the OpenWebText dataset\footnote{\url{https://huggingface.co/datasets/Skylion007/openwebtext}} consisting of 25k randomly selected sequences. The training configuration uses a batch size of 8, sequence length of 1024, one training epoch, and a learning rate of $5\times10^{-5}$.

The overall compression rate is set to 0.3. To achieve this target, only the MLP blocks are tensorized using Matrix Product Operators (MPO), as they contain the largest share of parameters and therefore offer greater potential for redundancy reduction. During the initial tensor decomposition stage, a uniform compression rate of 0.48 is applied to each selected layer.

In the local tensorization setting, each slice corresponds to an individual MLP block, which is optimized independently using the slice-wise feature distillation strategy described in Section~\ref{sec:meth}. In contrast, global tensorization performs conventional end-to-end finetuning over the full tensorized model.

For MPO construction, we use a two-site factorization, choosing balanced factorizations of the input and output dimensions whenever possible to avoid highly asymmetric MPO cores. The bond dimension is selected to match the target compression rate.

\subsubsection{Results}

We evaluate local and global tensorization on GPT-2 XL under the same optimization budget and compare both methods against the pretrained model. The goal of this experiment is not to establish a state-of-the-art LLM compression benchmark, but to assess whether the proposed local tensorization strategy remains effective and scalable in a large-model setting.

Table~\ref{table-gpt-comp} summarizes the results across multiple benchmarks. On LAMBADA, local tensorization outperforms global tensorization, achieving both lower perplexity (25.16 vs.\ 35.59) and higher accuracy (42.38\% vs.\ 35.51\%). This suggests that independently optimized feature-preserving slices can effectively recover task-relevant behavior in certain evaluation settings.

On WikiText and C4, however, global tensorization achieves lower perplexity than local tensorization, indicating that full end-to-end optimization remains advantageous for some large-scale language modeling benchmarks. Both tensorization strategies exhibit expected degradation relative to the pretrained model due to the limited tensorization dataset size. Further improvements may be achieved by scaling the training data and refining the optimization process.

Overall, these results suggest that local tensorization remains competitive with global tensorization while offering a fundamentally different optimization regime better suited to distributed execution.

\begin{table}[htbp]
\centering
\caption{Performance comparison of GPT-2 XL in terms of accuracy (acc, \%) and perplexity (ppl) across the pretrained dense model and tensorized variants.}
\label{table-gpt-comp}
\begin{tabularx}{\textwidth}{lXXX}
\toprule
\textbf{Benchmark} 
& \textbf{Dense GPT-2 XL} 
& \textbf{Tensorized Local} 
& \textbf{Tensorized Global} \\
\midrule
PIQA (acc)      & 70.51   & 61.32               & 61.70 \\
LAMBADA (ppl)   & 10.6341 & \textbf{25.1620}    & 35.5959 \\
LAMBADA (acc)   & 51.21   & \textbf{42.38}      & 35.51 \\
WikiText (ppl)  & 20.3766 & 45.5061             & \textbf{40.3350} \\
C4 (ppl)        & 50.0342 & 121.1180            & \textbf{100.6992} \\
\bottomrule
\end{tabularx}
\end{table}

Table~\ref{tab:time-gpt-comp} compares optimization time under local and global tensorization. We distinguish between single-GPU execution and an ideal distributed setting where local tensorization slices are optimized in parallel across multiple GPUs on different compute nodes.

In single-GPU settings, local tensorization can be slower than global tensorization because optimizing small independent slices leads to inefficient GPU utilization, as these workloads do not fully exploit parallel hardware capabilities. In contrast, global tensorization performs large dense computations over the full model, which are better suited to modern GPUs despite higher per-iteration cost. This effect is more pronounced for large models such as GPT-2 XL, where full-model training better leverages available compute resources. For smaller models like ResNet-34, however, the full model itself may not fully saturate the GPU, reducing the advantage of global optimization and allowing the lower computational overhead of local tensorization to make it more efficient.

In contrast, local tensorization naturally enables distributed parallel optimization because slices are independent and require no inter-slice gradient synchronization. Under an ideal parallel execution scenario where the 48 tensorized MLP slices are optimized concurrently, the total optimization time drops to 13.4 minutes.

This highlights the key systems-level trade-off of the proposed method: global tensorization is more efficient in single-device execution, whereas local tensorization is specifically advantageous when distributed parallel resources are available.

\begin{table}[htbp]
\centering
\caption{Optimization time comparison for GPT-2 XL. Parallel local tensorization corresponds to an ideal distributed execution scenario with independent slice optimization.}
\label{tab:time-gpt-comp}

\begin{tabular}{lcc}
\toprule
\textbf{Method}
& \makecell{\textbf{Parallel Time} \\ \textbf{(48 GPUs, min)}}
& \makecell{\textbf{Serial Time} \\ \textbf{(1 GPU, min)}} \\
\midrule
Local tensorization  & 13.4   & 531.35 \\
Global tensorization & --     & 110.25 \\
\bottomrule
\end{tabular}

\end{table}

\begin{figure}[!ht]
\centering

\begin{subfigure}[t]{0.7\textwidth}
    \centering
    \includegraphics[width=\linewidth]{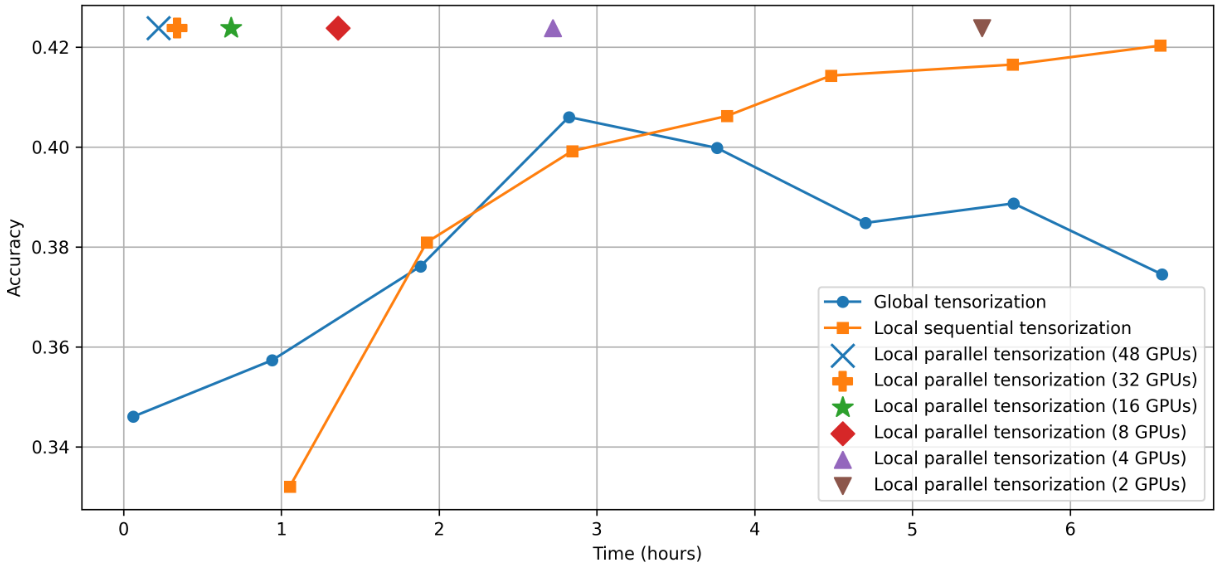}
    \caption{Test accuracy vs. optimization time}
    \label{fig:accuracy_time}
\end{subfigure}
\hfill
\begin{subfigure}[t]{0.7\textwidth}
    \centering
    \includegraphics[width=\linewidth]{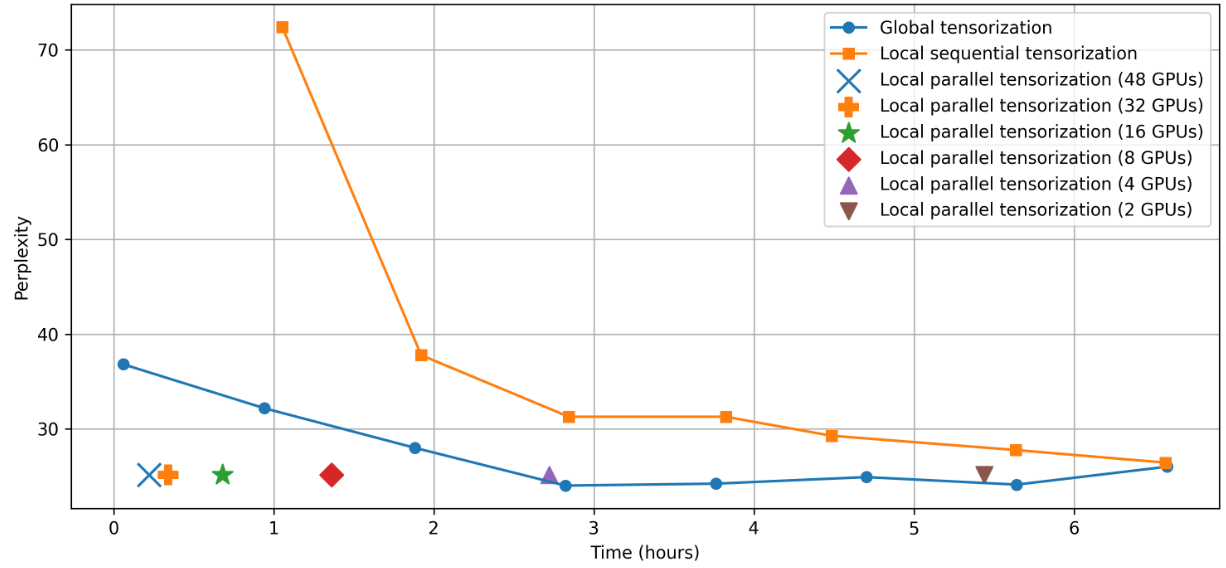}
    \caption{Perplexity vs. optimization time}
    \label{fig:perplexity_time}
\end{subfigure}

\caption{Performance of global and local tensorization methods as a function of optimization time on LAMBADA dataset. Local tensorization is shown in both sequential and parallel settings (2--48 GPUs).}
\label{fig:tensorization_time}
\end{figure}

Figure~\ref{fig:tensorization_time} illustrates performance as a function of optimization time. In the single-GPU setting, global tensorization improves more rapidly during early optimization, achieving stronger initial convergence in both accuracy and perplexity. The slight decrease in accuracy despite improving perplexity is expected, as these metrics capture different aspects of model performance: perplexity reflects overall token prediction quality, while exact-match accuracy is more sensitive to small prediction changes. Sequential local tensorization improves more gradually but remains competitive at later stages.

The parallel local tensorization results, shown as discrete markers corresponding to different levels of distributed execution (2--48 GPUs), demonstrate the main scalability advantage of the proposed approach. By exploiting slice independence, local tensorization reaches competitive performance substantially earlier in optimization time than either global tensorization or sequential local execution. Overall, these experiments show that while local tensorization is not inherently superior in single-device LLM optimization, it introduces a scalable optimization framework that becomes particularly attractive in distributed settings.

\section{Conclusions and perspectives}
\label{conclusions}

The proposed local tensorization strategy demonstrates strong modular scalability and efficient distributed parallelization without requiring inter-slice synchronization. While its slice-wise design may lead to lower hardware utilization in single-device settings, particularly for large models with relatively small slices, this presents a clear opportunity for improvement by grouping consecutive slices into larger optimization units. For example, two or more consecutive MLP blocks in a transfomer could be tensorized together via feature distillation of both the blocks treated as unit. Moreover, although local tensorization performs strongly at moderate compression rates, highly aggressive compression settings may benefit from a final global end-to-end finetuning step, as observed in the CIFAR-100 experiments. Future work will build on these promising results by extending the method to larger transformer architectures, exploring adaptive slice selection based on layer importance, investigating alternative loss functions such as cosine similarity, and combining tensorization with complementary compression techniques such as pruning and quantization.

\bibliographystyle{unsrt}
\bibliography{references}

\appendix

\section{Background: Tensorized Neural Networks}
\label{sec:background}

A tensorized neural network has at least one tensorized layer---a layer in which the weight matrix is represented as a tensor network using a specific tensorization algorithm (\cite{gao2020compressing}, \cite{liu2022deep},  \cite{zhang2017improved} \cite{xie2024neural},  \cite{ye2018learning}). This approach aims to reduce the number of parameters in the model while preserving model performance. In this paper, we employ two well-known tensorization algorithms to compress the weight matrices of a neural network: (1) Tucker decomposition and  (2) Matrix Product Operator (MPO)  decomposition. In this section, we briefly review these two methods before describing how to adapt them for local tensorization.

\subsection{Tensor networks}
A tensor network is composed of interconnected tensors, where a tensor is simply a multi-dimensional array, for example, an object of the class \texttt{numpy.ndarray} in Python. Each tensor has \textit{physical indices} that are left open or uncontracted and \textit{bond indices} that are contracted (summed over) with other tensors in the network (\cite{orus2014practical}). The size of the array dimension corresponding to a bond index is called the \textit{bond dimension} of that index. For example, consider the following contraction of a tensor network made of three tensors $A, B$ and $C$:

\begin{equation}
    F_{i_1,i_2, i_3}  = \sum_{j_1,j_2=1}^\chi A_{j_1,i_2}   B_{j_1,i_1,j_2}  C_{i_3,j_2}.
\end{equation}

Here, $\chi$ is the size of the two bond indices $j_1$ and $j_2$---assumed to be equal for simplicity, and $i_k$ for $k \in[1,3]$ are the physical indices which are left uncontracted. The result of this contraction is a tensor with physical indices of the three tensors.

In Python, a tensor network is a list of arrays attached to the nodes of a graph and the tensor network contraction is performed by, for example, \texttt{numpy.einsum()}\footnote{\url{https://numpy.org/doc/2.2/reference/generated/numpy.einsum.html}}. The \texttt{einsum()} string corresponds to the graph underlying the tensor network.

\subsubsection{Matrix Product operators (MPOs)}
\label{section_mpo}

Matrix Product operators, also called tensor train operators, are a popular example of tensor networks, which we employ in this paper to decompose dense linear layers inside a neural network. An MPO consists of $N$ four-index tensors arranged on a line. Each tensor has exactly two physical indices and either one or two bond indices for the endpoint and bulk tensors, respectively. For example, the MPO decomposition consisting of only $N$ tensors  is:

\begin{equation}
W_{i,j} = W_{(i_1,i_2,\dots,i_N),(j_1,j_2,\dots,j_N)}
=
\sum_{\chi_1,\chi_2,\dots,\chi_{N-1}}
T^{[1]}_{i_1,j_1,\chi_1}
T^{[2]}_{\chi_1,i_2,j_2,\chi_2}
\cdots
T^{[N]}_{\chi_{N-1},i_N,j_N}
\label{eq:mpo_general}
\end{equation}

where $i = i_1 \times i_2 \times \cdots \times i_N$ and
$j = j_1 \times j_2 \times \cdots \times j_N$ denote the Cartesian product
of the indices $i_1, i_2, \dots, i_N$ and $j_1, j_2, \dots, j_N$, respectively.
Here, $T^{[1]}$ and $T^{[N]}$ are the boundary tensors, while
$T^{[k]}$ for $2 \leq k \leq N-1$ are the bulk tensors, connected through
the virtual bond indices $\chi_1, \chi_2, \dots, \chi_{N-1}$.

\begin{figure}[htbp]
    \centering
    \includegraphics[scale=0.4]{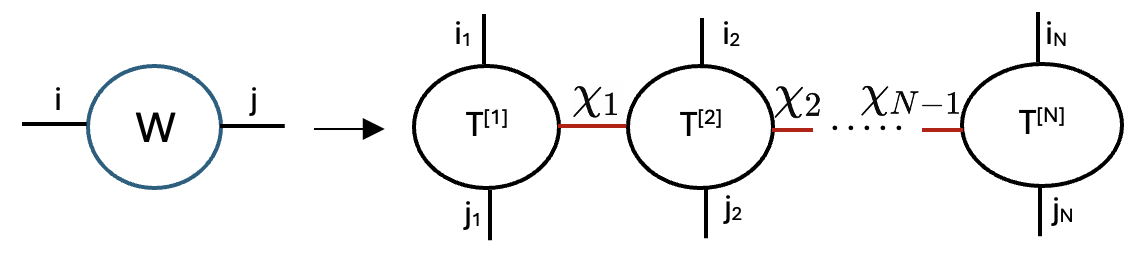}
    \caption{MPO decomposition of W into $N$ tensors}
\end{figure}

\begin{algorithm}
\caption{MPO Decomposition of a pretrained weight matrix W} \label{alg:mpo}
\begin{algorithmic}[1]
\STATE \textbf{Inputs:}
\\
(1) Weight matrix $W$, (2)  input indices $\{i_1, \dots, i_N\}$, (3) output indices $\{j_1, \dots, j_N\}$, (4) bond dimensions $\{|\chi_1|,|\chi_2|,\ldots,|\chi_N|\}$.
\STATE \textbf{Output:} MPO[1..N]: A list of truncated reshaped $U$ tensors.
\STATE \textbf{Initialize:} \STATE  $A = \operatorname{reshape}(W,(|i_1| |j_1| , \prod_{k=2}^N |i_k| |j_k|))$
\STATE $n = 2$
\WHILE{$n < N$}
    
    \STATE $U, \Sigma, V = \operatorname{Truncated SVD}(A)$
    \STATE $\mathsf{MPO}[n] = \operatorname{reshape}(U, (|\chi_{n-1}|, |i_n|, |j_n|, |\chi_{n}|) )$
    \STATE $A = \Sigma V$
    \STATE $n = n + 1$
\ENDWHILE
\STATE $\mathsf{MPO}[n] = \operatorname{reshape}(U, (|\chi_N|, |i_n|, |j_n|))$
\STATE \textbf{Return} MPO
\end{algorithmic}
\end{algorithm}

The process of MPO decomposition is given in Algorithm \ref{alg:mpo}. Let $W_{i,j}$ be the weight matrix of a linear layer with input and output dimensions $i$ and $j$, respectively. It is possible to obtain an MPO approximation of $W$ by iteratively applying truncated singular value decomposition to the weight matrix, $N-1$ times, where $N$ is the desired number of MPO cores. A weight matrix can be decomposed as follows:

\begin{equation}
    W_{i,j} = U_{i,i}\Sigma_{i,j} V_{j,j}^T
\end{equation}

Where $U$, and $V^T$  are the matrices of left and right singular values, respectively, and $\Sigma$ is a diagnoal matrix containing the singular values of $W$ in its diagonal sorted in ascending order. In order to obtain the MPO of  $W$, it is possible to define a number of  singular values $\chi$ and truncate the three matrices accordingly:
\begin{equation}
    W_{i',j'}^\chi = U_{i',\chi}\Sigma_{\chi,\chi} V_{\chi,j'}^T
    \label{svdk}
\end{equation}

In this first decomposition, $i'$ and $j'$ correspond to the new matrix indices after reshaping, such that $i'=i_1*j_1$, where $i_1$ and $j_1$ are the desired indices of the first MPO core, while $j'=\prod_{n=2}^{N} i_n j_n$, where $i_n$ and $j_n$ are the input and output dimensions of the  MPO cores left to create , respectively. 

As such, $U^{[1]}_{i,\chi}$ is the first tensor with one physical index of length $i$, and a contracted index with bond dimension $\chi$. This one can be reshaped into $U^{[1]}_{i_1,j_1\chi}$ to have a first MPO core. Then, it is possible to get a second core $U^{[2]}$ by contracting $\Sigma_{\chi,\chi}$ and  $V_{\chi,j}^T$ over the shared index $\chi$: $    U^{[2]}_{\chi,j} = \Sigma_{\chi,\chi}V_{\chi,j}^T$. $U^{[2]}_{\chi,j}$ can then be reshaped into $U^{[2]}_{\chi,i_2,j_2}$ to have a second MPO tensor with two physical indices.

Otherwise, if a higher number of cores is desired, then $U^{[2]}_{\chi,j}$ is reshaped and truncated SVD is applied on it as in equation \ref{svdk} to obtain a new tensor $U^{[n]}_{p,\chi}$, and so on. Here $n \in [1,N]$, and $N$ is the number of tensors in the network. 

Contrary to first and last MPO tensors, each intermediate tensor $U^{n}_{p,\chi}$ is reshaped into $U^{n}_{\chi,i_n,j_n,\chi}$, where $p = i_n \times j_n$. Here, we assume that the bond dimension $\chi$ is the same for all MPO cores, for simplicity.

Alternatively, one can directly initialize an MPO with the desired number of cores and bond dimensions such that $i = \prod_{n=1}^{N} i_n$, where $N$ is the desired number of cores and $i_n$ is the input dimension of the tensor $T^{[n]}$. The same applies to the output dimension $j = \prod_{n=1}^{N} j_n$, where $j_{n}$ is the output dimension of a tensor $U^{[n]}$.

In both cases, considering an MPO with N tensors, where $T^{[n]}$ is the $n$th tensor, the MPO decomposition of the  weight matrix $W$ can be expressed as:

\begin{equation}
\label{mpo:w}
    W_{i,j} = \sum_{\chi} \prod_{n=1}^{N} U^{[n]}_{\chi_{n-1}, i_n, j_n, \chi_n}
\end{equation}

Where $i =\prod_{n=1}^{N} i_n$, and $j = \prod_{n=1}^{N} j_n$, and $i_n$ and $j_n$ are input and output dimensions of the MPO tensors, respectively. The first contracted index $\chi_{n-1}$ of the first Tensor $U^{[1]}$, and last contracted index $\chi_{n}$ of the last tensor $U^{[N]}$, are set to 1. 

\paragraph{Deriving the bond dimension based on the compression rate for a 2-sites MPO:}

Given a targeted compression rate (the relative reduction in parameters) for a 2-d matrix, it is possible to derive the bond dimension for a 2-sites MPO as follows:

Let the matrix be factorized into two sites with input dimensions $(i_1, i_2)$ and output dimensions $(j_1, j_2)$, and define
$s_k = i_k j_k$. The number of parameters of the dense matrix is
\begin{equation}
P_{\mathrm{dense}} = (i_1 j_1)(i_2 j_2),
\end{equation}
while the MPO representation with bond dimension $\chi$ has
\begin{equation}
P_{\mathrm{MPO}} = i_1 j_1 \chi + i_2 j_2 \chi = (i_1 j_1 + i_2 j_2)\chi.
\end{equation}
Using the compression rate definition
\begin{equation}
\mathrm{CR} = 1 - \frac{P_{\mathrm{MPO}}}{P_{\mathrm{dense}}},
\end{equation}
we obtain
\begin{equation}
(i_1 j_1 + i_2 j_2)\chi = (1 - \mathrm{CR})(i_1 j_1)(i_2 j_2).
\end{equation}
Solving for the bond dimension gives
\begin{equation}
\chi = \frac{(1 - \mathrm{CR})(i_1 j_1)(i_2 j_2)}{i_1 j_1 + i_2 j_2}.
\end{equation}

\subsubsection{Tucker decomposition}

For a 2D convolution, Tucker decomposition can be applied to the corresponding 4D convolutional kernel as follows \cite{singh2024tensor}:

\begin{equation}
K_{x y i o} \approx \sum_{\alpha \beta \gamma \delta}
G_{\alpha \beta \gamma \delta}
U^X_{x\alpha}
U^Y_{y\beta}
U^{\mathrm{i}}_{i\gamma}
U^{\mathrm{o}}_{o\delta}.
\end{equation}

Here, the convolutional kernel \(K\) is factorized into a core tensor \(G\) and four factor matrices corresponding to the spatial and channel modes. The dimensions of the internal indices \( \alpha, \beta, \gamma, \delta \) define the Tucker ranks.

This decomposition reduces the number of parameters and computational complexity by approximating the original kernel with a compact representation while preserving its essential structure.

\begin{figure}[H]
    \centering
    \includegraphics[scale=0.4]{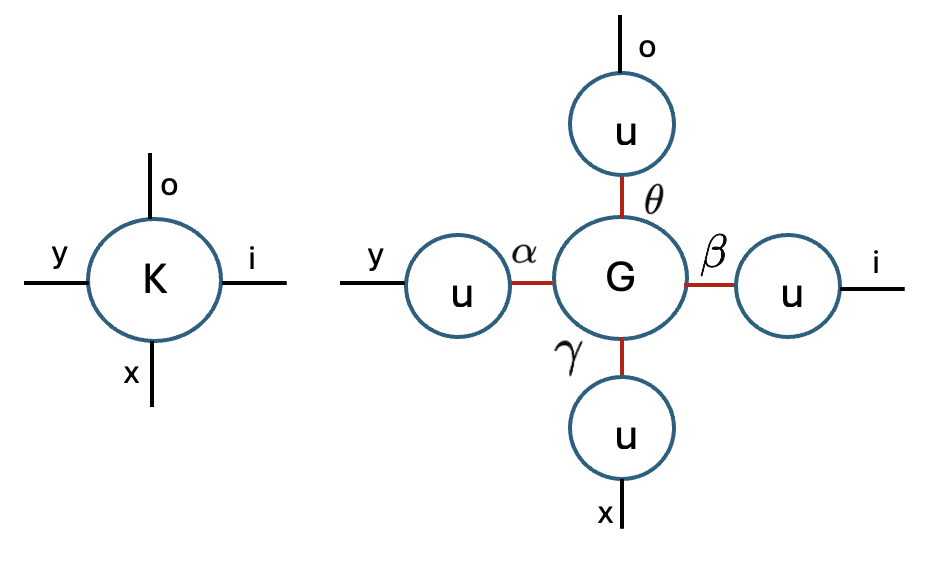}
    \caption{Tucker decomposition of a 4D convolutional kernel.}
\end{figure}

In practice, Tucker decomposition can be computed using Higher-Order SVD (HOSVD). The kernel tensor is unfolded along each mode:

\begin{equation}
K_{(n)} \in \mathbb{R}^{d_n \times \prod_{m \neq n} d_m}
\end{equation}

and truncated SVD is applied:

\begin{equation}
K_{(n)} \approx U^{(n)} \Sigma^{(n)} V^{(n)T}.
\end{equation}

The core tensor is then obtained by projection onto the reduced subspaces:

\begin{equation}
G =
K
\times_X (U^X)^T
\times_Y (U^Y)^T
\times_{\mathrm{in}} (U^{\mathrm{in}})^T
\times_{\mathrm{out}} (U^{\mathrm{out}})^T.
\end{equation}

The resulting compressed representation approximates the original convolution with fewer parameters and lower computational cost.

\section{Hyperparameter selection}

\subsection{Layer profiling results}

\vspace{-0.5cm}

\begin{figure}[H]
    \centering

    \begin{subfigure}{0.9\textwidth}
        \centering
        \includegraphics[width=\linewidth]{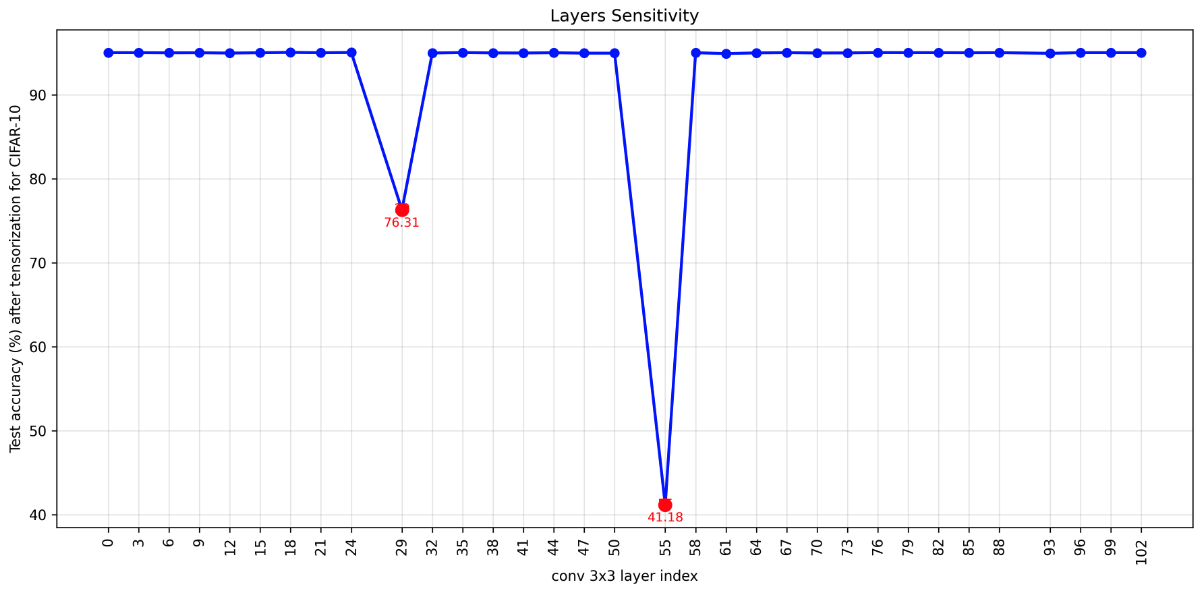}
        \caption{CIFAR-10}
    \end{subfigure}

    \vspace{0.3cm}

    \begin{subfigure}{0.9\textwidth}
        \centering
        \includegraphics[width=\linewidth]{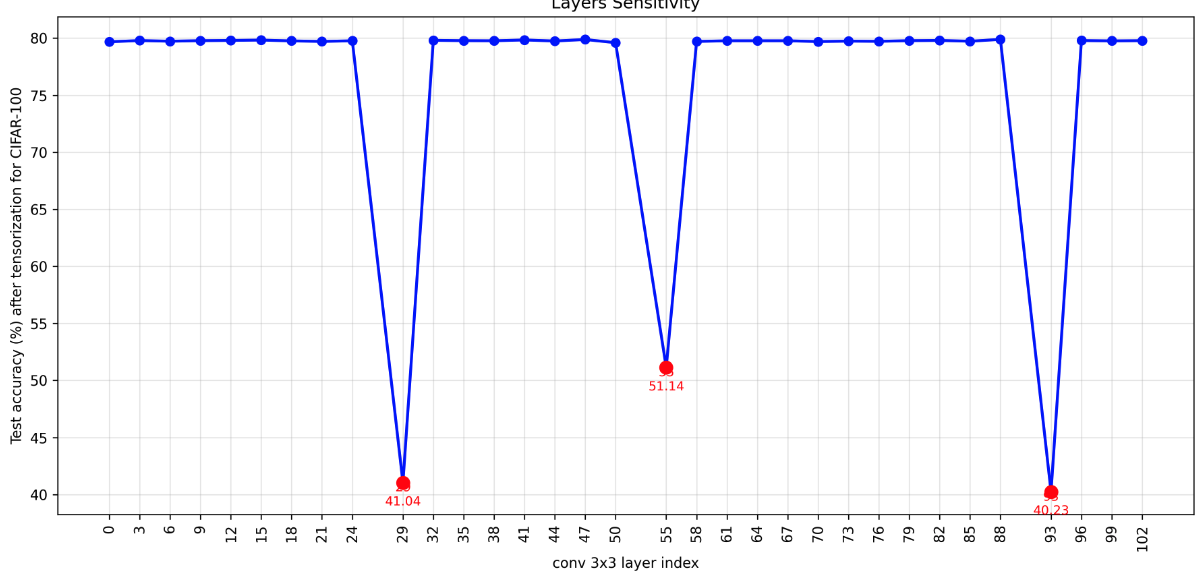}
        \caption{CIFAR-100}
    \end{subfigure}

    \caption{Sensitivity of individual layers with respect to test accuracy. For a convolutional layer with dimensions $[a,b,c,d]$, Tucker decomposition ranks are chosen as $[a/2,\, b/2,\, c,\, d]$, resulting in varying compression rates across layers.}
    \label{fig:layer-profiling}
\end{figure}

\subsection{Local training hyperparameters selection}

\begin{table}[H]
\centering
\caption{Effect of local training batch size on test accuracy after 10 epochs (best test accuracy reported, learning rate = 0.0005).}

\begin{subtable}[t]{0.46\textwidth}
    \centering
    \caption{CIFAR-10}
    \begin{tabular}{|c|c|c|c|c|c|}
        \hline
        Batch size & 8 & 16 & 32 & 64 & 128 \\
        \hline
        Test accuracy & \textbf{94.71} & 91.98 & 94.34 & 88.91 & 55.52 \\
        \hline
    \end{tabular}
\end{subtable}
\hfill
\begin{subtable}[t]{0.46\textwidth}
    \centering
    \caption{CIFAR-100}
    \begin{tabular}{|c|c|c|c|c|c|}
        \hline
        Batch size & 8 & 16 & 32 & 64 & 128 \\
        \hline
        Test accuracy & \textbf{78.89} & 77.19 & 77.89 & 71.12 & 19.93 \\
        \hline
    \end{tabular}
\end{subtable}

\label{tab:local-batch}
\end{table}

\begin{table}[H]
\centering
\caption{Effect of local training learning rate on test accuracy after 10 epochs (best test accuracy reported, batch size = 8).}

\begin{subtable}[t]{0.46\textwidth}
    \centering
    \caption{CIFAR-10}
    \begin{tabular}{|c|c|c|c|}
        \hline
        Learning rate & 0.0001 & 0.0005 & 0.001 \\
        \hline
        Test accuracy & 93.88 & 94.60 & \textbf{94.66} \\
        \hline
    \end{tabular}
\end{subtable}
\hfill
\begin{subtable}[t]{0.46\textwidth}
    \centering
    \caption{CIFAR-100}
    \begin{tabular}{|c|c|c|c|}
        \hline
        Learning rate & 0.0001 & 0.0005 & 0.001 \\
        \hline
        Test accuracy & 74.28 & \textbf{78.89} & 78.78 \\
        \hline
    \end{tabular}
\end{subtable}

\label{tab:local-lr}
\end{table}

\subsection{Global training hyperparameters selection}

\begin{table}[H]
\centering
\caption{Effect of global training batch size on test accuracy after 10 epochs (best test accuracy reported, learning rate = 0.0005).}

\begin{subtable}[t]{0.46\textwidth}
    \centering
    \caption{CIFAR-10}
    \begin{tabular}{|c|c|c|c|c|c|}
        \hline
        Batch size & 8 & 16 & 32 & 64 & 128 \\
        \hline
        Test accuracy & 81.81 & \textbf{84.86} & 84.09 & 83.69 & 81.58 \\
        \hline
    \end{tabular}
\end{subtable}
\hfill
\begin{subtable}[t]{0.46\textwidth}
    \centering
    \caption{CIFAR-100}
    \begin{tabular}{|c|c|c|c|c|c|}
        \hline
        Batch size & 8 & 16 & 32 & 64 & 128 \\
        \hline
        Test accuracy & 29.11 & 36.33 & 44.76 & \textbf{47.54} & 47.39 \\
        \hline
    \end{tabular}
\end{subtable}

\label{tab:global-batch}
\end{table}

\begin{table}[H]
\centering
\caption{Effect of global training learning rate on test accuracy after 10 epochs.}

\begin{subtable}[t]{0.46\textwidth}
    \centering
    \caption{CIFAR-10 (batch size = 16)}
    \begin{tabular}{|c|c|c|c|}
        \hline
        Learning rate & 0.0001 & 0.0005 & 0.001 \\
        \hline
        Test accuracy & 80.26 & \textbf{84.86} & 84.19 \\
        \hline
    \end{tabular}
\end{subtable}
\hfill
\begin{subtable}[t]{0.46\textwidth}
    \centering
    \caption{CIFAR-100 (batch size = 64)}
    \begin{tabular}{|c|c|c|c|}
        \hline
        Learning rate & 0.0001 & 0.0005 & 0.001 \\
        \hline
        Test accuracy & 37.46 & \textbf{47.54} & 40.55 \\
        \hline
    \end{tabular}
\end{subtable}

\label{tab:global-lr}
\end{table}

\end{document}